\documentclass{Interspeech2024}

\usepackage[nodisplayskipstretch]{setspace}
\usepackage{graphicx}
\usepackage{algorithm2e}
\usepackage{enumitem}
\usepackage{caption}

\title{AdaKD: Dynamic Knowledge Distillation of ASR models using Adaptive Loss Weighting}

\name[affiliation={*}]{Shreyan}{Ganguly}
\name[affiliation={*}]{Roshan}{Nayak}
\name[affiliation={*}]{Rakshith}{Rao}
\name[affiliation={*}]{Ujan}{Deb}
\name[affiliation={}]{Prathosh}{AP}

\address{Indian Institute of Science, India}
\email{gshreyan16@gmail.com, roshannayak610@gmail.com, rakshithdrao@gmail.com, ujandev@gmail.com, prathosh@iisc.ac.in}

\keywords{speech recognition, knowledge distillation, curriculum learning}

\begin{document}

\maketitle
\footnote{$^*$Equal contribution}

\begin{abstract}
    
    Knowledge distillation, a widely used model compression technique, works on the basis of transferring knowledge from a cumbersome teacher model to a lightweight student model. The technique involves jointly optimizing the task specific and knowledge distillation losses with a weight assigned to them. Despite these weights playing a crucial role in the performance of the distillation process, current methods provide equal weight to both losses, leading to suboptimal performance. In this paper, we propose Adaptive Knowledge Distillation, a novel technique inspired by curriculum learning to adaptively weigh the losses at instance level. This technique goes by the notion that sample difficulty increases with teacher loss. Our method follows a plug-and-play paradigm that can be applied on top of any task-specific and distillation objectives. Experiments show that our method performs better than conventional knowledge distillation method and existing instance-level loss functions. 
\end{abstract}

\section{Introduction}

In recent years automatic speech recognition (ASR) has seen radical improvements with the advent of deep neural network based models. Inspired by successful unsupervised pre-training approaches for text \cite{bert} and images \cite{doersch, henaff}, Schneider et. al. introduced wav2vec \cite{wav2vec} which trains a deep convolutional neural network with a noise contrastive objective on unlabeled data, improving upon previous baselines by upto 36\% \cite{wav2vec}. Baevski et. al. \cite{wav2vec2} further improved the state-of-the-art by pre-training a transformer encoder with quantized latent representations and a similar contrastive task. The widespread adoption of the transformer architecture across a wide range of modalities led to more transformer based ASR models such as HuBERT \cite{hubert} and data2vec \cite{data2vec}, all of which continue to push ASR performance to be more and more accurate. Even more recently, Radford et. al. introduced Whisper \cite{whisper}, a transformer trained on 680,000 hours of labeled data. In addition to matching the state of the art error rates on clean audio, Whisper far outperforms other models on zero shot, out of distribution generalization, advancing ASR robustness closer to human levels \cite{whisper}. Thus the current landscape in ASR is populated by many robust and powerful neural network models, which are getting closer and closer to human speech recognition capabilities.

A common use case of ASR models is real time transcription of speech. However despite their impressive performance, current neural network based ASR models are typically deep networks with parameters ranging from hundreds of millions to several billions \cite{wav2vec2, whisper}. This constrains their usage to devices with high compute resources. Large neural nets like large transformers also have a high inference latency, especially on CPU. Inference time for the whisper-medium model has been reported to be more than 60 secs for 30 sec audio inputs, even for high end CPUs like the AMD Ryzen 5 5600X \footnote{https://github.com/openai/whisper/discussions/454}. This prohibits real time usage for even the whisper-medium model, let alone larger more performant models like the whisper-large. This problem is further exacerbated when the models need to run on an edge device and cannot rely on remote compute over a network to do the transcription.

\begin{figure*}
\includegraphics[width=1.0\textwidth]{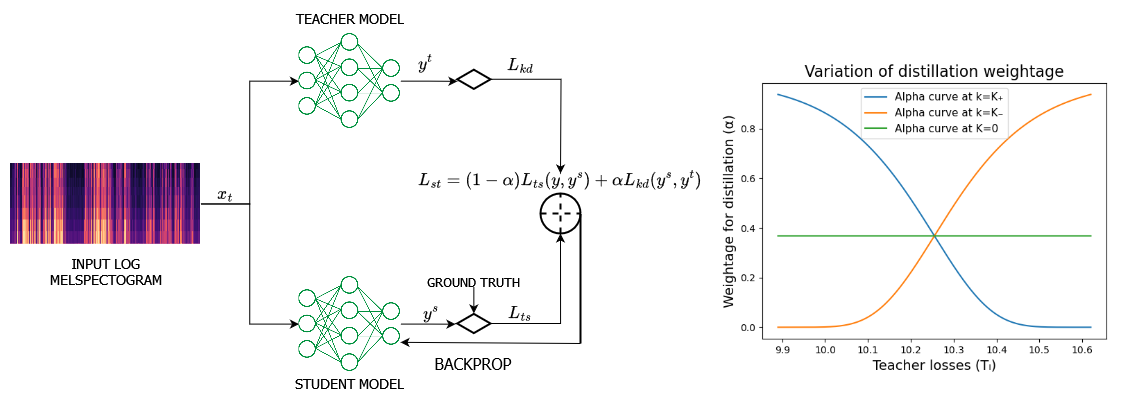}
\caption{\textnormal{A schematic diagram of the proposed knowledge distillation method with dynamic loss weights based on sample difficulty. The teacher model produces an output $y^{t}$ and incurs a loss $T_{l}$. The difficulty factor $d_{f}$ is calculated based on the teacher loss $T_{l}$ and two hyperparameters ($k$, $t$). The loss weight $\alpha$ for distillation is calculated using the difficulty factor $d_{f}$. The student model produces an output $y^{s}$ and incurs a loss $L_{ts}$. The final loss $L_{st}$ is a combination of the student loss $L_{ts}$ and the distillation loss $L_{kd}$.}}
\end{figure*} 

Therefore in order to run transformer based ASR models in real time we are prevented from directly using the large and powerful models. But large transformer models far outperform their smaller counterparts in terms of accuracy and robustness \cite{whisper}. A natural goal then is to attain the performance of a large model with a smaller and faster model. Model distillation aims to achieve this objective. Model distillation was formally introduced by Hinton et. al. \cite{hinton2015distilling} where the authors showed that task knowledge possessed by an ensemble of models can be distilled into a single smaller model, evidenced by its improvement on the task. In it's simplest case, the student model in addition to being trained on ground truth data is also asked to match the output logits of a teacher model, softened by a temperature hyperparameter. With respect to effective distillation approaches for transformers, Sanh et. al. introduced DistilBERT \cite{distilbert}, a distilled version of the BERT \cite{bert} model, where the authors performed knowledge distillation during pre-training, reducing model size and inference time by 40\% and 60\% respectively, while preserving performance of upto 97\% of the teacher BERT model \cite{bert}. Vision transformers \cite{vit} also saw successful distillation techniques applied to them \cite{tiny-vit}.

In speech recognition, Peng et. al. distilled knowledge from a wav2vec2 model \cite{wav2vec2} into smaller model with 4.8 times less parameters at the cost of about 7\% performance degradation. Fu et. al. and Kim et. al. further improved distillation on wav2vec2 by reducing model parameters by upto 91\% \cite{distilw2v2, kim22l_interspeech}. Lee et. al. applied knowledge distillation on the HuBERT \cite{hubert} model to obtain a model with 23.8\% less parameters with minimal performance degradation. The distillation approaches are a mostly a combination of logit and feature destillation and their variations. Peng et. al. apply logit distillation with a feature penalty \cite{bigfoot}. Fu et. al. adds a maximum mutual information (MMI) criterion on the layer representations \cite{distilw2v2}. To aid in compression, Kim et. al. share parameters and reuse attention alignments throughout the model \cite{kim22l_interspeech}.

While these different distillation approaches work well there is plenty of room for improvement. In particular, we notice that the distillation objectives employed above do not take into account the difficulty of the training samples, as encountered by the teacher. We show that this information can be captured from the teacher loss and utilized in order to improve knowledge distillation to the student model. To this end we propose a novel method where the distillation objective is dynamically adjusted throughout training so as to progressively shift knowledge distillation from easier to more difficult samples. We show consistent performance improvements across wav2vec2 \cite{wav2vec2} and whisper \cite{whisper} models and across 3 different languages.

\section{Related Works}
\textbf{Knowledge Distillation:} Leveraging the powerful generalization capacity of large cumbersome models to train smaller compact models with the aim of model compression was first proposed by Buciluǎ et. al. \cite{buciluǎ2006model}. This technique was extended in the work by Hinton et. al. \cite{hinton2015distilling} under the name of "Knowledge Distillation" using a new objective function that could be applied to both labeled and unlabeled sets. A brief description of knowledge distillation is provided in \autoref{section 3}. \par
The focus of recent works in this domain has been on improving the distillation performance by designing various forms of knowledge transfer schemes. It can be broadly categorized into logit distillation and feature distillation. In logit distillation, there have been works on choosing optimal teacher-student network architecture \cite{romero2014fitnets}, noise-based regularizer
for KD \cite{sau2016deep}, using conditional adversarial network to learn a loss function \cite{crowley2018moonshine}. Park et. al. \cite{park2019relational} proposed relational KD with the aim of transferring structural knowledge in logits from one model to another. Cho et. al. \cite{cho2019efficacy} provided a interpretation of the original KD, claiming a better and more accurate teacher does not ensure a more accurate student. 
\\
\textbf{Knowledge Distillation for ASR:} There has been works on compressing ASR models like Wav2Vec \cite{wav2vec} by Lee et. al. \cite{lee2022fithubert}, Peng et. al. \cite{peng2021shrinking}. However, for recent robust ASR models like Whisper \cite{whisper}, there has not been much advancement in the field. Gandhi et. al. \cite{shao2023whisper} used the vanilla knowledge distillation to distill a compressed version of a model from its base version. To this end, we try to present a novel KD-based approach for improving performance of Whisper model.

\textbf{Curriculum Learning:} Originally inspired by the human learning process, Curriculum Learning was introduced by Bengio et al. \cite{bengio2009curriculum} as a network training strategy involving the presentation of samples in a meaningful order. The recent trend in curriculum learning has predominantly focused on sequencing samples in increasing order of complexity during training. Castells et al. \cite{NEURIPS2020_2cfa8f9e} introduced a novel confidence-aware loss function for dynamic instance-wise curriculum learning. Wu et al. \cite{10167857} expanded on this approach by adopting a new scheme that separates difficult from incorrect samples within a curriculum learning framework. Recent literature also explores the application of curriculum learning in knowledge distillation. Li et al. \cite{li2023curriculum} and Zhu et al. \cite{zhu2021combining} proposed innovative knowledge distillation methods based on curriculum learning, progressing from easy to difficult samples using temperature and sample metrics. However, these studies have not delved into the exploration of incorporating increasing order of hidden knowledge difficulty, obtained while distilling, within the framework of curriculum learning.

\begin{table*}[th]
  \caption{A comaprison of Ada-Kd against other distillation methods. Whisper-small and Wav2Vec 2.0 XLS-R-1B finetuned on the mentioned datasets is the teacher and Whisper-tiny, Wav2Vec 2.0 XLS-R-300M pretrained as student. The value of hyperparameter $t$ taken here is the mean of teacher losses. Experiments for CV-Chinese and Aishell2 have not been performed for Wav2Vec 2.0 owing to poor teacher performance. CER metric has been used to compare the techniques.}
  \centering
  \begin{tabular}{*{9}{r} }
    \toprule
    \multicolumn{1}{c}{\textbf{Dataset/Model}} &
    \multicolumn{1}{c}{\textbf{Finetune}} &
    \multicolumn{1}{c}{\textbf{Normal KD \cite{hinton2015distilling}}} &
    \multicolumn{1}{c}{\textbf{Super Loss \cite{castells2020superloss}}} &
    \multicolumn{1}{c}{\textbf{Focal Loss \cite{tsung2017fl}}} &
    \multicolumn{1}{c}{\textbf{Annealing KD \cite{jafari2021annealing}}} &
    \multicolumn{1}{c}{\textbf{Adaptive KD}}\\                          
    \midrule
    $\text{CV-Hindi/Whisper}$~~~~ & 25.59 & 26 & 26.53 & 26.84 & 36.01 & \textbf{23.27}\\
    $\text{CV-Chinese/Whisper}$~~~   & 26.5 & 29.2 & 29.21 & 28.68 & 36.37 & \textbf{25.20}\\
    $\text{Aishell2/Whisper}$~~~ & \textbf{13.37} & 13.79 & 14.75 & 14.82 & 18.88 & 14.30\\
    $\text{CV-Tamil/Whisper}$~~~  & 13.39 & 13.12 & 12.32 & 14.31 & 13.59 & \textbf{12.07}\\
    $\text{CV-Hindi/Wav2Vec2}$~~~  & 17.52 & 15.04 & 16.63 & 15.78 & 25.59 & \textbf{14.26}\\
    \bottomrule
  \end{tabular}
  \vspace{5pt}

  \label{table:results}
\end{table*}
\section{Methodology} \label{section 3}
In certain cases, the standard loss functions fail to perform well on the task at hand and slight modifications on them will result in enhanced performance. In this section, we briefly touch upon the technique of knowledge distillation and then introduce the adaptive knowledge distillation technique aimed at instance-level adaptive loss weighting for knowledge distillation tasks. In addition, we try to delve into and elucidate the rationale behind the algorithm.

\subsection{Knowledge Distillation}
Model compression is a problem in which we aim to get the performance of a large model on a smaller one. Knowledge distillation is one such widely used model compression technique in which the large model, called the teacher is used as a guide while training another model smaller than its size called the student. In an ideal setting, the student will improve their performance by mimicking the teacher's behaviour. Hinton et. al. \cite{hinton2015distilling}, approached this problem by utilizing Kullback–Leibler divergence as the objective function to match the output logits of the teacher and student model as shown in Eq\eqref{equation:eq1}.

\vspace{-10px}

\begin{align}
  L_{kd}(y^{s},y^{t}) = \tau^{2}KL(y^{s},y^{t})
  \label{equation:eq1}
\end{align}

In addition, training can be done using both distillation and task-specific objectives to ensure that the student does not miss out on the particular goals of the task. Thus, the distillation objective comes out to be,

\vspace{-10px}

\begin{align}
  L_{student} = \sum_{i} (1 - \alpha)L_{ts}(y_{i}, x_{i}|\theta) + \alpha L_{kd}(y^{s}_{i},y^{t}_{i})
  \label{equation:eq2}
\end{align}

Here, the parameter $\alpha$ governs the significance of the distillation objective $L_{kd}$ in the training process. Moreover, the task-specific objective, denoted as $L_{ts}$, is assigned an importance level such that the combined importance of both objectives sum to one.

\subsection{Adaptive Knowledge Distillation}
The vanilla knowledge distillation method assigns equal weightage to task-specific and knowledge distillation losses. This does not capture the varying difficulty involved in transferring knowledge from teacher to student for different samples. In the past, instance-specific techniques \cite{tsung2017fl} \cite{castells2020superloss} have demonstrated promising results across a wide range of tasks. Inspired by curriculum learning \cite{bengio2009curriculum}, Adaptive Knowledge Distillation dynamically weighs the objectives for each instance based on its difficulty and the current training stage. We determine the difficulty of a sample based on the teacher loss, where the difficulty rises proportionally with it. We distill the knowledge from easier samples first, gradually increasing the distillation weightage for hard samples over the course of training. Moreover, we noticed that simultaneously decreasing the loss weights for easy samples assists in achieving optimal results. 

\vspace{-10px}

\begin{align}
  \alpha = e^{\frac{-1}{\sqrt{d_{f}}}}
  \label{equation:eq3}
\end{align}

To achieve this, we devised an equation \eqref{equation:eq3} that calculates the loss weight for distillation for each instance in a batch of 16. Where d is the difficulty factor and is calculated by the equation,

\vspace{-10px}

\begin{align}
  d_{f} = e^{-k(x-t)}
  \label{equation:eq4}
\end{align}

Here $d_{f}$ controlled by two hyperparameters, namely $k$ and $t$. The optimal values of these hyper-parameters depend on the distribution of the teacher losses on the training set used for distillation. The details of the values of these parameters have been discussed in section \ref{expts_and_results}. Ideally, $t$ can be the mean of teacher losses as shown below.

\vspace{-10px}

\begin{align}
  \text{Where}, t = \frac{1}{N}\sum_{i=1}^{N} L_{ts}(y_{i}, x_{i}|\theta)
  \label{equation:eq5}
\end{align}

Where N denotes the number of samples in the training dataset. While $k$ dictates the importance of distillation loss for each instance and is linearly decreased during the training. As a deliberate design choice, we set the initial value, $k_{+}$, such that the distillation loss weight for the sample with the highest teacher loss is a minimal value, specifically 0.1. Furthermore, $k_{-}$ is to be tuned. We conducted extensive experiments to find the optimal value for $k_{-}$. Figure 1 shows the alphas for samples based on teacher loss at two extreme values of $k$, i.e. $k_{+}$ and $k_{-}$ and also at $k=0$.







\begin{algorithm}
\SetAlgoNlRelativeSize{-1}

\vspace{0.5em}
\hrulefill \\[-1.2em]
\vspace{1.2em}
\textbf{Algorithm} Adaptive Knowledge Distillation \\
\hrulefill
\vspace{0.3em}

\textbf{Required:} Training data $X$, Validation data $V$, Student model initialization ${\theta}_s$, Teacher model ${\theta}_t$, Teacher losses $L_{teacher}$, task-specific loss $L_{ts}$, distillation loss $L_{kd}$, Range of hyperparameter $K$, Iteration steps $S$, learning rate $\eta$.
\vspace{0.5em}

\renewcommand\labelenumi{\theenumi:}
\begin{enumerate}[itemsep=0.01em]
    \item Initialize student model parameters ${\theta_s}$  
    \item Compute $L_{ts}(x, y|\theta_{t})$ for all $(x, y) \in X$ and choose $t$, $k_{+}$, and $k_{-}$.
    \item for $s \in \{0, 1, \ldots, S\}$ \textbf{do}
    \item \hspace{0.5cm} $x^\text{samples}, y^\text{samples} \xleftarrow{} MiniBatchSampler(X)$
    \item \hspace{0.5cm} Compute $\alpha$ for all $x^\text{samples}$ using Eq.~\eqref{equation:eq3}.
    \item \hspace{0.6cm} Form $L_{student}$ as given by Eq.~\eqref{equation:eq2}
    \item \hspace{0.5cm} Update $\theta_{s} \xrightarrow{} \theta_{s}-\eta \nabla L_{student}$  
    \item end for
\end{enumerate}



\end{algorithm}

\vspace{-10px}





\subsection{Rationale behind this algorithm} 
The soft probabilities from the teacher model play a crucial role in imparting insight into class relationships to the student. In instances where a robust teacher model consistently produces accurate predictions, a very small number of samples contribute to knowledge distillation. Furthermore, the vanilla knowledge distillation method assigns equal importance to all samples, failing to put emphasis on the harder ones to leverage the information about class relations. In contrast, our proposed method adopts a nuanced strategy by prioritizing the distillation of easier samples initially and progressively shifting focus to more challenging ones. This strategic shift is motivated by the understanding that the harder samples possess rich hidden knowledge, which conveys intricate class relationships, making them crucial for effective distillation once the knowledge from the easier samples has been distilled.


\section{Experiments and Results} \label{expts_and_results}

\subsection{Experimental setup}
\textbf{Toolkits.} Our experiments were implemented with the PyTorch \cite{paszke2019pytorch} and TorchAudio \cite{yang2021audio} deep learning frameworks. Pre-trained models were downloaded from HuggingFace \cite{wolf2019hf} repository.
\\
\textbf{Datasets.} Common Voice 11 and AiShell2 datasets were used for the experiments. Multiple languages were used, including Hindi, Tamil, and Mandarin, to test the robustness of our approach. AiShell2, an open source 1000 hours dataset that was split in a ratio of 80-20\% for train and test sets, respectively. 
\begin{center}
\captionof{table}{Datasets used and amount of train and test hours of validated data}
\resizebox{\columnwidth}{!}{
\begin{tabular}{|c|c|c|c|}
\hline
Dataset & Number of samples & Train hours & Test hours \\
\hline
CV-Hindi & 7.3K & 5 & 4 \\
\hline
CV-Tamil & 53k & 95 & 9 \\
\hline
CV-Chinese & 40k & 43 & 7 \\
\hline
Aishell2 & 250k & 800 & 200 \\
\hline
\end{tabular}
}
\end{center}

\textbf{Model.} We have used the variants of Whisper \cite{whisper} for our experiments, with Whisper-small as teacher and Whisper-tiny as student having 244 and 39 million parameters respectively. To verify the generalizability of our approach we further experimented on Wav2vec 2.0 \cite{wav2vec2}, with Wav2vec 2.0 XLS-R 1B as the teacher and Wav2vec 2.0 XLS-R 300M as the student.
\\
\textbf{Training.} The models were trained on Nvidia A100 GPU (80GB) with batch size of 16, learning rate of 1e-4 and Adam with the default initializations as optimizer. Cross-entropy was used as task specific loss and Kullback–Leibler divergence as distillation loss.

\subsection{Baselines}
We compare our methods with the existing instance-specific adaptive loss functions. Focal loss \cite{tsung2017fl} modifies the standard loss function in a way that reduces the loss of easy samples and, conversely, amplifies the same for hard samples. Thus, forcing the model to focus more on hard samples during the training. In contrast, super-loss \cite{castells2020superloss} calculates a confidence score of the model on each sample and downweights the contribution of samples to the loss with minimal confidence. Furthermore, annealing knowledge distillation \cite{jafari2021annealing} is aimed at progressively aligning the student model with the annealed soft probabilities of the teacher model during training. This is achieved by the introduction of a dynamic temperature term applied to the teacher model's soft probabilities.

\subsection{Results}
The results of the comparative study between our approach and the baseline techniques with a consistent experimental setup are presented in the Table \ref{table:results}. We have set the value of hyper-parameter $t$ to be the mean of teacher losses across all the experiments. It can be observed that the proposed technique outperforms both the language-specific fine-tuned model and vanilla knowledge distillation approaches across languages - solidifying its status to be a functional model compression technique. In case of dataset \textit{Aishell 2}, the performance of the fine-tuned model and vanilla distillation slightly outperforms the proposed technique, implying that with a very large data set, distillation methods cannot improve compared to fine-tuning. We compare our method against other baseline techniques. Super-loss, focal-loss, and annealing KD. For Annealing KD, we experimented with different $\tau$ hyperparameters in their formula to get the lowest CER. $\tau=4$ was the optimal choice for CV-Hindi whereas for other languages, $\tau=7$ performed the best for both the models. In terms of CER, it can be noted that our method outperforms these baselines considerably with the CER percentage 23. 27\%, 25. 20\%, 14. 30\%, 12. 07\% in the CV-Hindi, CV-Chinese, AiShell2, and CV-Tamil datasets, respectively. We have used $k$ ranges of 15 to -10 for Hindi, 8 to -8 for other whisper experiments. While the $k$ range for Wav2Vec 2.0 is observed to be very small of 0.024 to -0.015.

\subsection{Ablation Study}
In this study, we try to study how changing the hyperparameter $t$ will affect the performance of the distillation. In Table \ref{table:results} we had set $t$ to be the mean of teacher losses. We now run ablations by setting $t$ to the 25th, 50th, and 75th percentiles of the teacher losses and compare them against the mean. 

\vspace{-0.6em}
\begin{center}
  \includegraphics[width=0.8\linewidth]{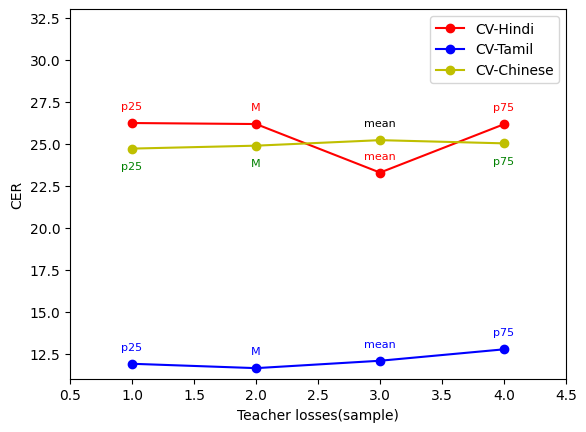}
\end{center}
\vspace{-0.6em}

\noindent We see that for CV-Hindi optimal results are obtained when $t$ is set to be the mean of the teacher losses. For CV-Chinese and CV-Tamil the mean does not come out to be the optimal choice. Even lower CERs are obtained for different $t$. These values perform close with an average deviation of 0.33 and 0.43 CER respectively from the CER of $t_{mean}$. This shows that, while $t_{mean}$ could be a good choice, depending on the distribution of the teacher losses, other values of $t$ could give optimal results.

\section{Conclusion}
In this work, we introduced a novel distillation technique to adaptively weigh the importance of losses at the instance level depending on teacher model performance. Inspired by curriculum learning, our method gradually transfers the relation between classes from the teacher to the student. The easy-to-use equation can be utilized on top of any task-specific and distillation loss function. Experimental results demonstrate that our method performs consistently better than existing instance-level loss functions on data sets of various sizes. We observe that our approach, like other distillation methods, falls short of standard fine-tuning if the dataset size is too large, as in case of AiShell2. Another potential shortcoming of our current approach is that $t$ and $k$ need to be tuned as hyperparameters. As future work we will look to address this by making these values learnable. We will also look to integrate this distillation technique with other efficiency approaches such as pruning and quantization.

\bibliographystyle{IEEEtran}
\bibliography{mybib}

\end{document}